\documentclass{article}
\usepackage{graphicx} 
\usepackage[table]{xcolor}
\usepackage{color, colortbl}

\usepackage{authblk}
\providecommand{\keywords}[1]{\textbf{\textit{Keywords:}} #1}
\usepackage{enumitem}

\usepackage{url} 
\usepackage{geometry}
\usepackage{longtable}
\usepackage{array}

\title{Quantum Algorithms: A New Frontier in Financial Crime Prevention} 

\author[1]{Abraham Itzhak Weinberg}
\author[2]{Alessio Faccia}
\affil[1]{AI-WEINBERG, AI Experts, Tel Aviv, Israel, aviw2010@gmail.com}
\affil[2]{University of Birmingham Dubai, Dubai, UAE, alessio.faccia@gmail.com}

\begin{document}
\maketitle

\begin{abstract}
Financial crimes’ fast proliferation and sophistication require novel approaches that provide robust and effective solutions. This paper explores the potential of quantum algorithms in combating financial crimes. It highlights the advantages of quantum computing by examining traditional and Machine Learning (ML) techniques alongside quantum approaches. The study showcases advanced methodologies such as Quantum Machine Learning (QML) and Quantum Artificial Intelligence (QAI) as powerful solutions for detecting and preventing financial crimes, including money laundering, financial crime detection, cryptocurrency attacks, and market manipulation. These quantum approaches leverage the inherent computational capabilities of quantum computers to overcome limitations faced by classical methods. Furthermore, the paper illustrates how quantum computing can support enhanced financial risk management analysis. Financial institutions can improve their ability to identify and mitigate risks, leading to more robust risk management strategies by exploiting the quantum advantage. This research underscores the transformative impact of quantum algorithms on financial risk management. By embracing quantum technologies, organisations can enhance their capabilities to combat evolving threats and ensure the integrity and stability of financial systems.
\end{abstract}


\section{Introduction}
Entering the Quantum Decade, as highlighted by IBM \cite{IBM}, signifies the beginning of a transformative period in quantum computing. This decade is expected to see the development of quantum computing from theoretical exploration to practical applications that tackle some of the most complex challenges across various industries. IBM’s Quantum Network, a global community of over 210 Fortune 500 companies, leading academic institutions, startups, and national research labs, illustrates a collective effort to utilise the potential of quantum computing. This network exemplifies collaborative efforts to accelerate progress in quantum computing, with the ultimate aim of developing applications that solve previously unsolvable problems. As we explore the capabilities of quantum computing, especially in improving quantum algorithms for fighting financial crimes, it is clear that the Quantum Decade is not just a prediction of technological progress but a call to action for organisations to prepare for the quantum revolution. \\
Quantum Machine Learning (QML) strives to develop and implement techniques that can be executed on quantum computers to tackle the standard tasks of supervised, unsupervised, and Reinforcement Learning (RL) found in classical Machine Learning (ML). QML’s utilisation of quantum operations sets it apart, coupling the extraordinary properties of quantum computing, such as superposition, tunnelling, entanglement, and quantum parallelism. This paper relates to Quantum Neural Networks (QNNs), the quantum equivalent of classical neural networks. \\
QML represents a complex intersection of quantum computing and machine learning (ML), aiming to revolutionise how we approach data analysis, pattern recognition, and decision-making processes. QML addresses the core tasks of supervised, unsupervised, and reinforcement learning central to classical ML but with a distinctive twist by leveraging techniques executable on quantum computers. Integrating quantum operations—taking advantage of superposition, tunnelling, entanglement, and quantum parallelism—enables QML to process information in ways that classical systems cannot match. At the heart of this innovative field are Quantum Neural Networks (QNNs), which embody the quantum counterpart to classical neural networks.
QNNs promise to enhance computational capabilities significantly \cite{mangini2021quantum}, offering new avenues for complex problem-solving and algorithmic efficiency. Unlike their classical counterparts, QNNs can theoretically handle vast datasets with unparalleled speed, thanks to their ability to operate in high-dimensional Hilbert spaces. This capability is particularly promising for tasks requiring large-scale data analysis in bioinformatics, climate modelling, and financial forecasting. Furthermore, QML is not merely about transposing existing ML algorithms into the quantum domain; it includes creating new algorithms native to quantum computing. These quantum-specific algorithms are designed to exploit the natural advantages of quantum states, potentially leading to algorithmic complexity and computational speed breakthroughs. For instance, quantum algorithms for pattern recognition can operate more efficiently by exploring all possible patterns simultaneously due to quantum superposition.\\
Additionally, QML’s exploration of quantum entanglement offers novel ways to represent and understand complex correlations in data that classical algorithms struggle to untie \cite{ray2024quantum}. This aspect is particularly intriguing for developing more sophisticated models of human cognition and social networks, where the intricacies of interconnections and influences are beyond the reach of classical computation. The synergy between quantum computing and machine learning, as embodied in QML and QNNs, heralds a new era of computational intelligence. By transcending the limitations of traditional computing paradigms, QML opens the door to solving some of the most challenging problems facing various scientific and industrial fields today. As we stand on the brink of this computational revolution, the potential applications and implications of QML and QNNs continue to unfold, promising profound impacts on technology, society, and our understanding of systemic complexity. \\
Table \ref{table:FCCATEG} below categorises the most relevant types of Financial Crimes. It offers a comprehensive framework for classifying financial crimes, integrating and harmonising key concepts from globally recognised entities such as the Association of Certified Fraud Examiners (ACFE), the Financial Action Task Force (FATF), Anti-Money Laundering (AML) standards, the United Nations Office on Drugs and Crime (UNODC), and the International Monetary Fund (IMF). This framework encapsulates the vast range of criminal activities these organisations aim to combat by categorising financial crimes into broad areas such as fraud and scams, money laundering, terrorist financing, corruption and bribery, cybercrime, market abuse, tax evasion, and identity theft. Each category and subcategory reflects the definitions and concerns highlighted by these entities, ensuring a broad coverage that includes various types of financial crimes identified in their reports and recommendations.\\
The framework’s inclusivity allows for a structured approach to understanding financial crimes, offering a solid foundation for further analysis, policy formulation, and the development of countermeasures. For instance, the FATF’s focus on money laundering and terrorist financing finds a direct correspondence in Table \ref{table:countermeasures}, ensuring that the global priorities in combating financial crimes are well-represented. Similarly, the ACFE’s extensive work on fraud and corruption is mirrored in the detailed subcategories under ‘Fraud and Scams’ and ‘Corruption and Bribery’, respecting the depth of their research and guidance. The inclusion of cybercrime acknowledges the growing concern of entities like the UNODC and IMF regarding the digital dimension of financial crimes, emphasising the evolving nature of these threats.
Moreover, the framework’s structure facilitates easy mapping and integration of specific guidelines, recommendations, and reporting requirements from these organisations, making it a versatile tool for academic and practical applications. It supports the alignment of national and international efforts in the fight against financial crimes, encouraging a unified approach that can enhance effectiveness and cooperation across borders.\\
In essence, Table \ref{table:FCCATEG} serves as a foundational piece in the larger puzzle of global financial crime prevention, offering a starting point to explore, match, and integrate the diverse and rich perspectives of leading organisations in the field. It aims to foster a holistic understanding and approach to combating financial crimes, reflecting the complexity and interconnectivity of these illegal activities in the modern world.

\begin{table}
  \centering
    \begin{tabular}{ |p{3.8cm}|p{4.5cm}|p{6.5cm}| }
    \hline
    \rowcolor{blue!30}
    \textbf{Category} & \textbf{Subcategory} & \textbf{Description} \\
    \hline
    \hline
    Fraud and "Scams" & Consumer Fraud & Deceptive practices causing loss in consumer transactions. \\
     & Corporate Fraud & Misconduct within corporations for financial gain. \\
     & Securities and Investment Fraud & Market manipulation, Ponzi schemes, insider trading. \\
     & Insurance Fraud & False claims to obtain insurance benefits. \\
    \hline
    Money Laundering and Proceeds of Crime & Traditional Money Laundering & Concealing origins of illegally obtained money through transactions. \\
     & Cryptocurrency Laundering & Using digital currencies to hide the source of funds. \\
     & Asset Laundering & Reinvesting criminal proceeds into assets to conceal origins. \\
    \hline
    Terrorist Financing & Domestic Financing & Funding terrorist activities within a country. \\
     & International Financing & Transferring funds across borders to support terrorism. \\
    \hline
    Corruption and Bribery & Public Sector Corruption & Misuse of official powers in the public sector for personal gain. \\
     & Private Sector Corruption & Bribery and misconduct in private industry. \\
     & Transnational Bribery & Bribes paid to foreign officials for business advantages. \\
    \hline
    Cybercrime & Financial Data Breaches & Unauthorized access to financial data leading to theft or fraud. \\
     & Online Fraud and Scams & Cyber methods targeting individuals or organizations for financial gain. \\
     & Ransomware and Extortion & Demanding payment to prevent or undo cyberattacks. \\
    \hline
    Market Abuse and Insider Trading & Market Manipulation & Artificially inflating or deflating the price of securities. \\
     & Insider Trading & Trading based on non-public, material information. \\
    \hline
    Tax Evasion and Avoidance & Domestic Tax Evasion & Illegally hiding income from national tax authorities. \\
     & Offshore Tax Evasion & Using foreign jurisdictions to evade taxes. \\
     & Aggressive Tax Planning & Exploiting loopholes to minimize taxes without direct illegality. \\
    \hline
    Identity Theft and Financial Impersonation & Identity Theft & Using someone else's personal information for financial gain. \\
     & Financial Impersonation & Assuming a false identity to commit financial crimes. \\
    \hline
  \end{tabular}
  \caption{Categories of Financial Crimes.}
  \label{table:FCCATEG}
\end{table}

\clearpage
\begin{longtable}{|p{3.5cm}|p{3.5cm}|p{3.5cm}|p{5.5cm}|} 
\hline
\rowcolor{blue!30}
\textbf{Financial Crimes} & \textbf{Technological Countermeasures} & \textbf{ML Countermeasure Approach} & \textbf{Examples of Quantum Algorithms for Countermeasure Approaches} \\
\hline
\hline
Consumer Fraud & AI-powered fraud detection systems & Supervised learning for pattern recognition & Quantum anomaly detection algorithms use Quantum Support Vector Machines (QSVM) to identify unusual patterns. \\
\hline
Corporate Fraud & Digital forensic tools & Network analysis for detecting fraudulent transactions & Grover’s search algorithm for fast retrieval enhances quantum clustering algorithms to identify groups with similar fraudulent patterns. \\
\hline
Securities and Investment Fraud & Blockchain for transparent transactions & Anomaly detection algorithms & Quantum Principal Component Analysis (QPCA) for dimensionality reduction in large datasets, improving classification accuracy. \\
\hline
Insurance Fraud & Predictive analytics for claim analysis & Classification algorithms for fraudulent claims & Quantum algorithms for pattern recognition and anomaly detection in claims data, potentially using Quantum Approximate Optimisation Algorithm (QAOA). \\
\hline
Traditional Money Laundering & AI-powered transaction monitoring & Clustering algorithms for detecting suspicious activities & Quantum machine learning (QML) for network analysis, using quantum-enhanced clustering to reveal hidden structures in transaction data. \\
\hline
Cryptocurrency Laundering & Blockchain analytics tools & Anomaly detection in cryptocurrency transactions & Quantum simulations to trace complex cryptocurrency flows and identify laundering schemes with unprecedented speed and accuracy. \\
\hline
Asset Laundering & Digital asset tracking systems & Machine learning models for asset tracing & Quantum-enhanced algorithms for analysing asset registration and transaction patterns to uncover illicit financial flows. \\
\hline
Domestic Financing & Network analysis tools & Regression analysis for fund tracking & Quantum algorithms for complex network simulations, aiding in the identification of covert funding channels for terrorist activities. \\
\hline
International Financing & Cross-border transaction monitoring & Deep learning for pattern recognition in financial flows & Quantum simulations to model and analyse international financial networks for tracking terrorist financing across borders. \\
\hline
Public Sector Corruption & E-governance platforms & Text mining for anomaly detection in procurement data & Quantum computing techniques for deep analysis of government procurement data to identify patterns indicative of corruption. \\
\hline
Private Sector Corruption & Compliance monitoring AI & Predictive analytics for risk assessment & Quantum-enhanced pattern recognition in corporate transactions and communications to detect bribery and misconduct. \\
\hline
Transnational Bribery & International database integration & Machine learning for identifying risk patterns & Quantum algorithms for analysing global financial transactions to uncover transnational bribery schemes. \\
\hline
Financial Data Breaches & Encryption technologies & Anomaly detection for unauthorised access & Quantum key distribution (QKD) for secure communications, ensuring data integrity against cyber threats. \\
\hline
Online Fraud and Scams & Behavioural analytics & Supervised learning for email filtering & Quantum-enhanced machine learning for real-time fraud detection, significantly reducing false positives in online transactions. \\
\hline
Ransomware and Extortion & Anti-malware tools with AI & Predictive models for threat detection & Quantum-resistant encryption methods to safeguard data against ransomware attacks and ensure privacy. \\
\hline
Market Manipulation & Real-time market surveillance systems & Time series analysis for detecting manipulation & Quantum optimisation algorithms to solve complex market prediction models more efficiently than classical algorithms. \\
\hline
Insider Trading & Anomaly detection in trade data & Classification models for unusual trading patterns & Quantum algorithms for analysing vast amounts of market data and communications to detect insider trading activities. \\
\hline
Domestic Tax Evasion & Automated tax reporting and analysis systems & Regression models for income prediction & Quantum Fourier Transform (QFT) algorithms identify cyclical patterns in financial data indicative of evasion schemes. \\
\hline
Offshore Tax Evasion & International tax compliance databases & Network analysis for offshore entities & Quantum-enhanced analysis of global financial transfers to detect and trace offshore tax evasion strategies. \\
\hline
Aggressive Tax Planning & AI for legal document analysis & Machine learning for loophole identification & Quantum computing is used to analyse complex tax planning strategies and identify potentially abusive schemes. \\
\hline
Identity Theft & Biometric verification systems & Machine learning for biometric authentication & Quantum-enhanced machine learning for real-time identity verification, significantly reducing the risk of identity theft. \\
\hline
Financial Impersonation & Digital identity verification platforms & Deep learning for face recognition & Quantum simulations for detecting and analysing patterns of financial impersonation across digital platforms. \\
\hline
\caption{Technological and Algorithmic Countermeasures to Financial Crimes.}
\label{table:countermeasures}
\end{longtable}

The above classification matches the technological countermeasures, the ML Countermeasure Approaches and the Examples of Quantum Algorithms that can be used for each case as it is disclosed in Table \ref{table:countermeasures}.

\section{Advantages of Quantum Artificial Intelligence (QAI) Over Conventional AI Methods}
Quantum Artificial Intelligence (QAI) offers several advantages over conventional Artificial Intelligence (AI) methods \cite{dunjko2018machine}. Quantum algorithms provide exponential speedups for ML, simulation, and optimisation tasks, enabling real-time or predictive analysis of significantly larger datasets. Additionally, QAI can solve problems that are intractable for classical computers, such as integer factorisation and certain machine learning tasks, due to the exponential resource requirements of these problems. Moreover, the massive parallelism of quantum systems allows QAI models to potentially recognise complex patterns across vast amounts of data, surpassing the capabilities of classical AI. Quantum techniques like amplitude amplification and quantum dimensionality reduction excel at automatically discovering meaningful features without human input. Quantum representations can encode highly non-linear and intricate relationships between variables, often present in real-world data.
Furthermore, entanglements of quantum systems have the potential to provide new intuitive insights about data beyond what classical models offer. Certain QAI algorithms, like variational circuits, can maintain advantages even in near-term device noise.\\ Additionally, combining quantum and classical techniques into hybrid systems may yield capabilities greater than either paradigm alone. QAI offers performance, representational, and insightful benefits compared to conventional AI methods. Its speed, scalability, pattern recognition abilities, feature discovery techniques, representation of non-linear relationships, potential for new insights, noise tolerance, and hybrid approaches all contribute to its promise in advancing AI capabilities.
In addition to the above, several metrics can be used for comparing QAI to traditional ML, such as Accuracy, Training time, Inference time, Data size, Scalability, Generalization, Flexibility, Explainability, Energy efficiency, and Cost of training/deployment \cite{gill2022ai, wang2020quantum}.\\
Firstly, in terms of accuracy, QAI can potentially achieve higher accuracy with larger and more complex datasets, thanks to quantum advantages. Secondly, QAI models could significantly reduce training time by leveraging quantum acceleration or amplification. Additionally, QAI may enable near-instant inference on large datasets, thereby improving inference time. Moreover, QAI has the potential to effectively process massive datasets that surpass the limits of classical computing, addressing big data problems.
Furthermore, quantum parallelism could exponentially enhance the scalability of QAI models, allowing for the incorporation of more training data and parameters. QAI models may also exhibit better generalisation by enhancing pattern recognition capabilities. With quantum computational flexibility, QAI models can be easily adapted or retrained for new tasks, enabling highly versatile lifelong learning models. In terms of explainability, certain QAI approaches, such as variational circuits, may offer more interpretability compared to deep learning models. Additionally, QAI’s inherent parallelism can lead to significantly lower power consumption, making it more energy-efficient for specific tasks. Finally, while there may be initial costs associated with hardware, expertise, and infrastructure, the cost of training and deploying QAI models is expected to decrease over time as technology progresses, making it a more economical choice.

\section{Unleashing the Power of Quantum Computing in Financial Crime}
Quantum computing offers a range of powerful applications in financial crime detection and prevention. QML models, such as Quantum Neural Network (QNN) and Quantum Support Vector Machine (QSVM), have the potential to analyse massive financial transaction datasets much faster than classical AI methods \cite{innan2023financial}. They can identify complex money laundering or fraud schemes that might go unnoticed by conventional approaches. Quantum Natural Language Processing (QNLP) can swiftly sift through various texts, including reports and calls, to detect signs of economic crimes by better understanding semantics and contextual concepts \cite{guarasci2022quantum, trivedi2021loan}. \\
Quantum Reinforcement Learning (QRL) agents can interact with synthetic environments that simulate financial systems at scale \cite{niu2019universal, meyer2022survey}. It enables them to identify new vulnerabilities exploited by criminals, providing valuable insights for strengthening security measures. Quantum recommendation systems can learn customer behaviours and promptly detect anomalous activities \cite{kerenidis2016quantum}, allowing for immediate flagging and investigation before significant losses occur.
Furthermore, quantum dimensionality reduction techniques extract meaningful features from high-dimensional time series and other real-world financial data. It enhances the ability to accurately detect anomalies and suspicious patterns \cite{cong2016quantum}. Quantum clustering algorithms also play a significant role by grouping related criminal activities or entities that transcend borders or employ dissimilar techniques \cite{aimeur2007quantum}, providing a more comprehensive understanding of interconnected criminal networks.\\
Quantum simulations of markets and investment strategies offer insights into systemic risks and allow for testing interventions at a speed and scale that classical computers cannot match \cite{daley2022practical}. Finally, combining the advantages of both paradigms, hybrid quantum-classical models can potentially transform how financial intelligence agencies and banks leverage data to prevent financial harm. By harnessing the power of quantum computing alongside classical methods \cite{de2022survey}, these hybrid models can revolutionise the approach to combating financial crimes. In summary, the applications of quantum computing in financial crime detection and prevention are vast and promising. From analysing large datasets to understanding complex semantics and detecting anomalies in real-time, quantum computing is poised to reshape the context of safeguarding financial systems and preventing illicit activities.\\
As mentioned before, QML algorithms, such as QSVM, can revolutionise the analysis of large transaction datasets by detecting patterns that indicate money laundering, fraud, and other illicit activities \cite{nosal2023crime,girasa2022regulation}. The quantum version of these algorithms can perform the classification task exponentially faster. In addition, quantum annealing and Variational Quantum Eigensolver (VQE) algorithms offer the opportunity to optimise risk models for extensive portfolios, enabling the detection of abnormal financial activities associated with economic crimes like insider trading. Quantum Principal Component Analysis (QPCA) and other dimension reduction techniques can extract meaningful features from multidimensional financial time series, aiding in identifying anomalies and suspicious transactions related to theft and embezzlement.
Furthermore, quantum order finding algorithms, such as Shor's algorithm \cite{childs2003quantum}, can analyze periodic patterns in financial market indicators and transactions, uncovering cyclical behaviour exploited by perpetrators of economic crimes such as pump-and-dump schemes. By leveraging quantum graph algorithms, particularly quantum walk-based algorithms applied to transaction networks, hidden connections between entities involved in economic crimes can be detected faster than with classical methods \cite{kalra2018portfolio}. Quantum simulations of financial processes and models have the potential to accelerate stress testing of controls and risk analysis, facilitating the identification of new vulnerabilities exploited by economic criminals \cite{dalzell2023quantum}. Moreover, quantum sampling abilities can generate realistic synthetic economic crime datasets, which can be used to train machine learning models for detection and prevention. Finally, combining hybrid quantum-classical algorithms with investigative techniques and domain expertise holds the promise of revolutionising economic crime fighting. By harnessing the power of quantum computing alongside classical approaches, we can potentially enhance our ability to combat and prevent economic crimes more effectively.\\
Financial crimes such as money laundering, terrorist financing, fraud, and tax evasion have significant financial implications for society, costing billions of dollars each year. These illicit activities erode trust in the financial system and facilitate more severe criminal behaviours. In the past, financial institutions primarily relied on human analysts to monitor transactions and report suspicious activities. However, modern financial systems’ scale, complexity, and speed have surpassed human capabilities. Enormous volumes of data are generated daily, holding the potential to reveal criminal patterns. Advanced analytics and AI offer promising solutions to this challenge. By processing vast amounts of unstructured data, including transactions, documents, and communications, AI technologies can augment human analysts and identify intricate patterns and anomalies that might go unnoticed. Equipped with the right tools and access to aggregated transaction data, regulators and law enforcement agencies can more effectively detect criminal networks, track illicit financial flows, and prevent unlawful funds from infiltrating the financial system.\\ From a business perspective, financial firms aim to fortify their platforms against criminal exploitation, minimise fines and reputational risks, and assure customers that their assets are secure. Advanced monitoring and AI-based systems safeguard the interests of all stakeholders involved. Moreover, the application of these new technologies holds great promise in enabling authorities to swiftly and efficiently identify threats to public safety, such as terrorism, human trafficking, corruption, and other forms of organised criminal schemes that undermine society as a whole.\\
QML models, such as QNN and QSVM, offer the ability to analyse massive financial transaction datasets at a much faster pace than classical AI methods. These models excel in identifying complex money laundering or fraud schemes that may elude traditional approaches. Quantum natural language processing enables quick sifting through texts like reports and calls, leveraging a better understanding of semantics and contextual concepts to detect signs of economic crimes more effectively \cite{grossi2022mixed, omolara2018state}. Furthermore, quantum reinforcement learning agents can interact with synthetic environments that simulate financial systems on a large scale. It allows them to identify new vulnerabilities exploited by criminals, providing valuable insights for enhancing security measures. Quantum recommendation systems learn customer behaviours and promptly detect anomalous activities, enabling immediate flagging for investigation, thus mitigating potential significant losses.\\
Quantum dimensionality reduction techniques extract meaningful features from high-dimensional time series and other real-world financial data, improving anomaly detection capabilities. Quantum clustering algorithms comprehensively group related criminal activities or entities that transcend borders or employ dissimilar techniques, enhancing the understanding of interconnected criminal networks. Moreover, quantum simulations of markets and investment strategies contribute to a deeper comprehension of systemic risks and enable testing interventions at a speed and scale not achievable with classical computers. Finally, hybrid quantum-classical models, harnessing both paradigms’ advantages, can fundamentally transform how financial intelligence agencies and banks leverage data to prevent financial harm. In summary, quantum computing applications in financial crime detection and prevention encompass many capabilities. From analysing massive datasets and understanding complex semantics to identifying vulnerabilities and mitigating risks, quantum computing has the potential to revolutionise the way we safeguard financial systems and combat illicit activities.\\
The effective detection of financial crimes relies on the utilisation of advanced analytics, AI, and extensive financial data to uncover both familiar and novel manipulation techniques employed by criminal organisations. Transaction monitoring plays a vital role in identifying suspicious patterns, including atypical fund transfers, deposits/withdrawals, and mingling of personal and business funds, as well as structuring tactics to evade reporting thresholds. Customer due diligence is essential during onboarding to establish risk profiles, while ongoing monitoring ensures that these profiles remain consistent with transactional activities. Inconsistencies and suspicious identification documents serve as red flags. Entity resolution enables linking interconnected accounts, entities, and individuals that may be intentionally disguised, thereby exposing larger schemes and criminal networks.\\
Network analysis facilitates the mapping of connections between transacting parties by exploring common accounts, addresses, and phone numbers, revealing clusters and intermediaries indicative of underground banking operations. Fund flow analysis traces the trajectory of funds over time through intricate networks, shedding light on the funds’ ultimate source, destination, and purpose, thereby unravelling layers of transactions designed to obfuscate illicit activities. Behavioural analysis involves constructing models of normal customer behaviours and establishing transaction benchmarks, enabling the identification of significant deviations such as changes in transaction locations or volumes that deviate from established patterns and triggering alerts. Predictive modelling leverages financial and contextual data to develop risk scores and prioritise alerts, empowering investigators to focus on higher-risk potential leads. Finally, data aggregation from internal and external sources provides valuable context concerning customers and geopolitical risks, facilitating comprehensive risk assessments. By harnessing the power of advanced analytics, AI, and extensive data, financial institutions and law enforcement agencies enhance their ability to detect and combat financial crimes, both well-known and innovative, perpetrated by criminal organisations.\\
QML algorithms, such as QSVM and quantum Boltzmann machines, offer powerful tools for analysing patterns in large databases to identify anomalies indicative of crimes like fraud, embezzlement, and insider trading \cite{egger2020quantum, guo2022quantum}. These algorithms can process vast amounts of data and detect subtle deviations more efficiently than classical methods by leveraging the computational advantages of quantum computing. Additionally, quantum graph algorithms, specifically quantum walks applied to networks of entities and transactions, can potentially uncover hidden connections between perpetrators of complex financial crimes at an accelerated pace \cite{kadian2021quantum}. It enables the detection of intricate relationships and networks that may not be readily apparent using conventional approaches \cite{innan2023financial}. Quantum clustering algorithms \cite{ramezani2020machine} provide an automated means of grouping related financial crime activities involving different entities and accounts across various locations and time frames. By identifying common patterns and associations, these algorithms contribute a more comprehensive understanding of criminal activities.\\
Quantum Order Finding (QOF) and period detection algorithms, such as Shor’s algorithm, analyse periodic transaction behaviours, allowing for the identification of activities such as Ponzi schemes and pump-and-dump operations. It enables timely detection and intervention in fraudulent practices. Quantum dimension reduction techniques, such as PCA and Singular Value Decomposition (SVD), extract meaningful financial features that can be used to detect outliers and flag suspicious activities for further investigation. These techniques improve anomaly detection capabilities and enhance the accuracy of crime identification. Quantum state and process tomography enable rigorous testing of controls and the identification of weaknesses that financial cybercriminals may exploit \cite{white2022non}. By conducting simulations and analysing the behaviour of financial systems, these techniques provide valuable insights for strengthening security measures.\\
Hybrid quantum-classical optimisation approaches can accelerate risk modelling of large portfolios and databases, enabling the detection of abnormal flows that may indicate financial crimes. By leveraging both quantum and classical computing capabilities, these models improve the efficiency and accuracy of risk assessment. Quantum generative models offer the ability to expand limited financial crime datasets synthetically. By generating realistic data, these models facilitate the robust training of classical and QML algorithms, improving their performance in identifying and preventing financial crimes. Furthermore, quantum analogue and digital simulation techniques aid in modelling the dynamics of financial systems at scale. It provides a deeper understanding of macro patterns exploited by organised crime syndicates, aiding in developing more effective prevention and intervention strategies. In summary, quantum algorithms’ speed, scale, and parallelism significantly enhance the capabilities of detecting diverse financial crimes. By leveraging quantum machine learning, graph algorithms, dimension reduction techniques, order finding, state tomography, hybrid optimisation, generative models, and simulation methods, financial institutions can strengthen their defences against criminal activities and protect the integrity of financial systems.

\section{Quantum Computing for Malicious Finance Detection: Benefits and Evaluation Metrics}
Quantum computing offers significant benefits for the detection of malicious finance activities. First, quantum algorithms provide exponential speedups, enabling near real-time analysis of huge datasets impractical for classical computers \cite{naik2023portfolio}. Additionally, quantum computing efficiently handles datasets and systems of a scale that exceeds the capabilities of classical approaches \cite{njorbuenwu2019survey}. Furthermore, quantum approaches can potentially expose complex correlations and patterns hidden in classical methods, enhancing detection accuracy \cite{arslan2018study}. The speed of quantum computing also enables interactive exploration of edge cases and facilitates what-if analysis, offering insights that are not feasible to obtain classically. Moreover, hybrid quantum-classical tools may provide valuable insights into the origins and mechanics of detected activities, contributing to their explainability.\\
Various metrics can be considered to evaluate the effectiveness of quantum-based malicious finance detection properly. One important metric is the time to detection, which quantifies the reduction in the time taken to identify malicious activities using quantum analysis compared to classical methods. Another metric is the detection rate, which measures the percentage of simulated or historical attempts that quantum models successfully flag compared to classical approaches. Striking a balance between detection coverage and minimising unnecessary alerts is crucial, and this can be assessed through the false positive rate metric. Additionally, evaluating the impact of additional or refined data fields and features on detection performance, known as feature sensitivity, helps optimise the detection process. The robustness of the models against adversarial manipulation of data or strategies aimed at avoiding detection is another important metric to consider.\\
Scalability metrics focus on determining the maximum volume of data, entities, and interconnections that can be effectively analysed within a given timeframe using quantum computing. Furthermore, the fidelity and interpretability of explanations provided by quantum models can be quantitatively assessed to aid in investigating detected malicious activities. Finally, the capability of quantum computing to identify previously unknown activities or schemes beyond what classical methods can achieve, known as novelty, is an essential metric to consider. Proper evaluation against these benefits and metrics is crucial to validate the value proposition of quantum computing in the context of detecting and combating malicious finance activities. It enables a comprehensive assessment of the effectiveness, efficiency, and practical applicability of quantum approaches, ultimately supporting informed decision-making regarding the adoption of quantum solutions in this domain.

\section{Enhancing Money Laundering Detection with Quantum Algorithms}
The application of financial analytics, network analysis, and machine learning plays a crucial role in detecting patterns, relationships, and risk factors that may go unnoticed by humans but could potentially indicate money laundering or other forms of financial crime. Various techniques are employed for this purpose, including transaction monitoring, customer due diligence, funds flow analysis, behavioural analysis, network analysis, transaction benchmarking, transaction structuring detection, predictive modelling, entity resolution, and data aggregation.\\
Transaction monitoring involves scrutinising wire transfers, deposits, withdrawals, and other activities for suspicious patterns, including cross-border flows. Customer due diligence entails screening customers against watchlists and sanction lists and monitoring transactions to ensure consistency with customer risk profiles. Fund flow analysis focuses on tracing the movement of funds across interconnected accounts, corporations, and individuals to unveil money laundering schemes. Behavioural analysis utilises machine learning to establish profiles of typical account usage and identify anomalous transactions that deviate from established behaviours. Network analysis aims to map connections between transacting parties, identifying clusters, common intermediaries, and underground banking systems. \\ Transaction benchmarking involves comparing transaction amounts, frequencies, geographies, and counterparties against peer groups to flag outliers. Transaction structuring detection aims to identify the intentional fragmentation of transactions to evade reporting requirements. Predictive modelling involves developing models that assess transaction risks and prioritise alerts for investigators. Entity resolution focuses on linking related transactions, accounts, and identities that may be intentionally disguised. Finally, data aggregation combines internal and external data sources to gain a more comprehensive view of risks. By leveraging these techniques and technologies, financial institutions and law enforcement agencies can enhance their ability to combat financial crime effectively.\\
As mentioned before, QML algorithms, such as QSVMs, can analyse large transaction datasets and swiftly identify suspicious patterns indicative of money laundering. These quantum algorithms outperform their classical counterparts, enabling faster and more accurate classification tasks. Additionally, quantum graph algorithms, specifically quantum walks applied to transaction networks, excel in detecting hidden connections between entities and uncovering complex money laundering rings or schemes at an accelerated pace compared to classical algorithms \cite{mishra2024enhancing, grossi2022mixed}. \\
Quantum clustering algorithms are invaluable in automatically grouping related money laundering activities based on transaction attributes, revealing connections that may not be apparent through traditional means. Quantum dimensionality reduction techniques, such as QPCA, extract meaningful features from high-dimensional financial data. It facilitates the detection of anomalies and the flagging of suspicious transactions for further investigation. Furthermore, the quantum order finding algorithm can analyse periodic or cyclical patterns in financial activities, allowing for the identification of behaviour typical of specific money laundering techniques like round-tripping or smurfing. Quantum state tomography, applied to the output of quantum simulations of financial networks or processes, enables stress testing of controls and the identification of vulnerabilities exploited by money launderers.\\
Hybrid quantum-classical optimisation algorithms can potentially accelerate risk modelling of large and complex portfolios, enabling the detection of abnormal flows that may indicate money laundering activities. Additionally, quantum sampling techniques can aid in generating realistic synthetic money laundering datasets, augmenting the training of machine learning models and improving their effectiveness. Indeed, the speed and scalability of quantum algorithms significantly enhance money laundering detection capabilities. By leveraging quantum machine learning, graph algorithms, dimensionality reduction techniques, order finding algorithms, state tomography, hybrid optimisation algorithms, and sampling methods, financial institutions can more effectively combat money laundering and safeguard their systems against illicit activities.

\section{Enhancing Cryptocurrency Crime Detection with Quantum Computing}
The detection and prevention of financial crimes employ advanced algorithms and leveraging expanding financial crime databases to triangulate structural and behavioural analyses. Screening transactions against watchlists of known criminals, sanctioned entities, and politically exposed persons is a significant measure to identify potentially corrupted interactions. Analysing transaction metadata, including payee names, memo fields, and account descriptors, provides valuable context for categorising transactions and uncovering illicit schemes.\\
Geotagging techniques play a crucial role in identifying abnormal transaction locations that may indicate fraudulent activity while monitoring cross-border flows, which assists in detecting potential movements of illicit funds. Structural analysis delves into examining relationships, communication patterns, and fund movements within networks to expose hierarchies, connections, and the division of labour within criminal organisations. \\
Link and cluster analysis identifies associations between accounts and entities over time, drawing upon attributes such as overlapping addresses, phone numbers, and IP addresses, exposing hidden relationships. Machine learning-based anomaly and outlier detection utilise historical databases of past criminal and non-criminal transactions and patterns to identify statistical deviations that necessitate further investigation automatically. A enhances and expedites these techniques by continuously learning from investigators, adapting models, and recommending new monitoring patterns as criminal methods evolve.Effective coordination and information sharing between the public and private sectors are also essential in combating financial criminal networks. This collaborative approach ensures the efficient exchange of intelligence and maximises the collective effort to combat financial crimes. \\
As mentioned above, QML algorithms, particularly QSVM, offer powerful tools for analysing patterns in graphs of cryptocurrency transactions and addresses \cite{girasa2022regulation}. These algorithms can swiftly flag suspicious entities and activities indicative of theft, scamming, and money laundering. By leveraging the computational advantages of quantum computing, these algorithms outperform classical methods, enabling faster detection of illicit activities in the crypto space \cite{ganapathy2021quantum}. \\
Quantum period finding algorithms contribute to the identification of repetitive patterns in cryptocurrency transfers, which often signify pump-and-dump schemes or round-tripping activities associated with market manipulation crimes. This capability assists in the timely detection and prevention of fraudulent practices \cite{gomes2023fortifying}. Quantum clustering algorithms excel in grouping crypto transactions and entities involved in complex crimes that span multiple accounts and currencies, overcoming the limitations of classical clustering techniques. This comprehensive approach enhances understanding of interconnected criminal networks within the cryptocurrency ecosystem. Quantum signature detection techniques analyse various attributes of crypto transactions, such as coin amounts and velocities, to identify anomalies and outlier transactions associated with theft and fraud crimes \cite{guo2022quantum}. By leveraging quantum computing’s computational power, these algorithms can efficiently detect suspicious activities that may go unnoticed by traditional methods.\\
Quantum dimension reduction methods, such as PCA and SVD, extract meaningful financial features from high-volume and complex cryptocurrency transaction data. It enables the identification of outliers and flags transactions for further investigation, enhancing anomaly detection capabilities. Cryptocurrency network analysis, utilising quantum walk algorithms applied to networks of entities and transactions, uncover hidden connections between players involved in syndicated crypto crimes. By revealing intricate relationships, these algorithms provide valuable insights for combating organised criminal activities within the crypto space. Quantum simulation techniques allow for modelling the dynamics of large cryptocurrency markets and systems at a scale that classical computers cannot match. It provides a deeper understanding of macro patterns exploited by organised criminal networks, aiding in developing effective prevention and intervention strategies.\\
Hybrid quantum-classical tools integrate the results obtained from quantum analytics with traditional forensics, enabling improved detection of cryptocurrency crimes. By combining the strengths of both paradigms, these tools enhance the accuracy and efficiency of crime detection and investigation in the crypto domain. In summary, the parallelism and scalability of quantum computing significantly enhance the capabilities of detecting and combating cryptocurrency crimes. Through the utilisation of quantum machine learning, period finding, clustering, signature detection, dimension reduction, network analysis, simulation, and hybrid quantum-classical tools, the cryptocurrency ecosystem can be safeguarded against illicit activities, protecting investors and maintaining the integrity of the market.

\section{Enhancing Detection of Market Manipulation with Quantum Computing}
Unravelling manipulative schemes that are often intentionally designed to evade detection requires the implementation of sophisticated algorithms, pattern-matching techniques, and entity resolution. Manipulative practices such as identifying wash trades must be addressed, which involve fake transactions to create misleading activity or liquidity in a stock without changing beneficial ownership. Monitoring for spoofing is essential, which entails entering large orders and cancelling them to artificially influence price movements before executing trades. Additionally, detecting pump-and-dump schemes is crucial, whereby coordinated buying artificially inflates the price of a stock, followed by selling to dispose of shares on unsuspecting retail investors.\\ Analysing the concentration of trades is necessary to identify instances where one entity dominates the trading volume in a stock, particularly around news announcements or filing periods. Comparing trade volumes to the public float is important, as abnormally high volumes could indicate the trading of fake shares. Benchmarking price moves against news events is also crucial, as drastic price changes without corresponding news raise concerns.
The behavioural analysis of traders plays a significant role in identifying coordinated patterns among entities and using the same brokerages or accounts. \\
Assessing order-to-trade ratios is essential, as high proportions of cancelled orders compared to completed trades could signal spoofing. Network analysis of trading relationships helps identify dense clusters of entities repeatedly engaging in trades with each other. Machine learning techniques can be employed, using past cases to train models to detect newer manipulation techniques autonomously. Ongoing monitoring and the adaptability of models are vital to keep pace with the evolving tactics employed by manipulators. By leveraging these sophisticated tools and techniques, it becomes possible to detect and mitigate manipulative activities in financial markets effectively.\\
We have already mentioned that QML algorithms, such as QSVM, offer powerful tools for analysing patterns in large datasets of stock trades, orders, and prices \cite{saxena2023financial,egger2020quantum}. By leveraging these algorithms, market manipulation strategies like pump-and-dump and spoofing can be automatically identified faster. The computational advantages of quantum computing enable efficient processing of vast amounts of data, improving the detection of unfair practices in the stock market. Quantum graph analysis techniques, particularly quantum walks applied to trading entities and transaction networks, can uncover complex insider trading rings and hidden coordination between actors. By examining the relationships and interactions within the market, these algorithms provide valuable insights into illicit activities that may not be easily detectable using traditional methods \cite{peelam2023quantum}. \\
Quantum dimension reduction methods, such as Quantum Principal component analysis (QPCA) and Qunatum SVD (QSVD), extract meaningful features from high-dimensional market data. It enables the identification of anomalous trades, prices, or volatility patterns that may indicate market manipulation. Through quantum computing’s computational power, these techniques enhance the accuracy of identifying suspicious activities.\\
Quantum signal processing techniques focus on detecting engineered price or volume signals that could indicate front-running or insider trading activities. By analysing stock market signals in the quantum domain, these algorithms can identify patterns that may not be apparent using classical methods, contributing to detecting unfair practices in real-time. Quantum order-finding algorithms analyse repetitive or cyclical trading periodicities, which may wash trading or other strategies designed to influence markets artificially. By leveraging quantum computing’s capabilities, these algorithms provide insights into trading patterns that can assist in exposing market manipulation techniques. \\
Quantum portfolio optimisation techniques allow stress tests on large and complex portfolio holdings using quantum solvers. It enables verifying compliance with regulations, particularly regarding conflicts of interest, enhancing transparency and accountability in the financial industry. Quantum simulation techniques enable the modelling of stock exchange dynamics and the co-movement of related firms’ prices at scale. By gaining insights into potential market manipulation techniques, these simulations contribute to a deeper understanding of unfair practices in financial markets.

\section{Quantum Circuit Born Machines in Finance: Enhancing Generative Modeling and Quantitative Applications}
A Quantum Circuit Born Machine (QCBM) is a QML model based on Born machines and offers several key advantages \cite{alcazar2020classical}. Born machines are variational hybrid quantum-classical models that leverage Born’s rule from quantum mechanics to evaluate classical data distributions. In the case of QCBMs, the Born rule distribution is implemented using a parameterised quantum circuit, enabling the model to harness the power of quantum hardware \cite{ganguly2023implementing}.\\
The quantum circuit encodes a quantum state representing the data distribution, and measurements in this state provide samples from the distribution. The circuit parameters are optimised variationally using a classical optimiser to minimise the difference between the generated and target distributions. This approach offers a quantum-enhanced generative modelling capability that can efficiently learn certain distributions compared to classical methods. QCBMs have demonstrated their potential in various tasks, including density estimation, molecular simulations, graph neural networks, and generative adversarial networks. One of the advantages of QCBMs is that they have relatively weaker resource requirements than other QML models, making them implementable on near-term quantum devices. However, it is important to note that their practical implementation is still constrained by the size of circuits that can be realised.\\
In finance, QCBMs hold promise for various applications \cite{coyle2021quantum}. For instance, in option pricing and hedging, QCBMs can model complex underlying asset distributions more accurately than classical methods, enabling the pricing of exotic options and effective risk management of large option portfolios. \\
QCBMs can also aid in portfolio optimisation by capturing the intricate covariances between assets and providing recommendations for optimal asset allocations. Risk modelling becomes more refined with QCBMs as they can simulate extreme but plausible scenarios from the learned distribution, assisting in estimating portfolio Value at Risk and Expected Shortfall. Furthermore, QCBMs can be utilised for anomaly detection in finance, helping to identify outliers and anomalies in time series data, such as stock prices, by learning normal distributions and flagging deviations. In fraud detection, QCBMs can model complex patterns in large financial transaction graphs, enabling the identification of traces of money laundering, spoofing, and pump-and-dump schemes. Macroeconomic forecasting can benefit from QCBMs as they can capture complex non-linear relationships in economic indicators, facilitating predictions of recessionary environments and commodity prices.
Stress testing of banks’ resilience to market crashes and liquidity crises can be enhanced through QCBMs by simulating distributions of risk factor shocks.\\
In trading strategies, QCBMs can aid in developing and evaluating strategies like market making by modelling order books and limiting order placements. Additionally, QCBMs can be used for synthetic data generation, producing realistic training datasets by sampling from QCBM distributions to augment classical models. The unique modelling abilities offered by QCBMs have the potential to improve various quantitative finance applications significantly. By leveraging the power of quantum computing, QCBMs can provide more accurate and efficient solutions for pricing, risk management, anomaly detection, fraud detection, forecasting, stress testing, trading strategies, and synthetic data generation in the financial industry.

\section{Advancing Risk Management in Finance: Quantum Computing for Enhanced Analysis}
Quantum computing holds immense potential for revolutionising risk management in the financial industry. By leveraging the speed, scale, and parallelism inherent in quantum systems, various applications can be developed to enhance risk analysis and decision-making. Here are some key areas quantum computing can significantly impact \cite{how2023business,woerner2019quantum}.\\
Quantum portfolio optimisation allows using quantum optimisation techniques, such as VQE \cite{tilly2022variational}, to efficiently determine optimal asset allocations for large and complex portfolios while considering risk constraints. This approach has the potential to outperform classical methods and deliver superior portfolio performance. Quantum scenario analysis introduces quantum simulations to rigorously assess risks under extreme market conditions that are challenging to model using classical approaches. By leveraging the computational power of quantum systems, financial institutions can gain deeper insights into the potential impact of rare and catastrophic events on their portfolios.\\
Quantum Value-at-Risk (VaR) estimation involves leveraging QML algorithms, like QSVMs, to estimate VaR and Expected Shortfall (ES) risk measures using vast historical market data. Quantum algorithms can process and analyse this data more efficiently, enabling more accurate and timely risk assessments. Quantum risk factor modelling employs QPCA techniques to extract risk factors from high-dimensional market time series data. Financial institutions can improve risk modelling and management strategies by identifying the underlying factors driving portfolio risk.\\
Quantum correlation analysis offers the ability to analyse the joint interactions of multiple risk factors exponentially faster through quantum algorithms. This enhanced capability enables a more comprehensive and accurate portfolio risk modelling, particularly in complex and interconnected markets. Quantum counterparty risk analysis involves applying quantum graph algorithms to map connections between financial entities, identifying potential systemic risks and vulnerabilities within the financial system.\\
By leveraging quantum computing’s ability to handle and analyse large-scale networks, financial institutions and regulators can gain deeper insights into counterparty risk and systemic stability.
Quantum stress testing utilises hybrid quantum-classical optimisation approaches to test the resilience of risk models under diverse assumption shocks at a large scale. By combining the computational power of quantum systems with classical optimisation techniques, financial institutions can assess the robustness of their risk management frameworks and identify areas for improvement. \\
Quantum anomaly detection leverages QML techniques to flag outlier market moves within massive datasets. By harnessing the power of quantum systems, financial institutions can enhance their risk investigation capabilities and promptly address potential anomalies and irregularities. Overall, the speed, scale, and parallelism offered by quantum computing have the potential to advance risk management capabilities for financial institutions and regulators significantly. By embracing quantum technologies and developing innovative applications, the financial industry can enhance risk analysis, improve decision-making processes, and strengthen overall financial stability.

\section{Comparing Quantum Algorithms and Classical Approaches: Key Findings}
\begin{figure}[!ht]
	\begin{center}
		\includegraphics
		[scale=0.3]{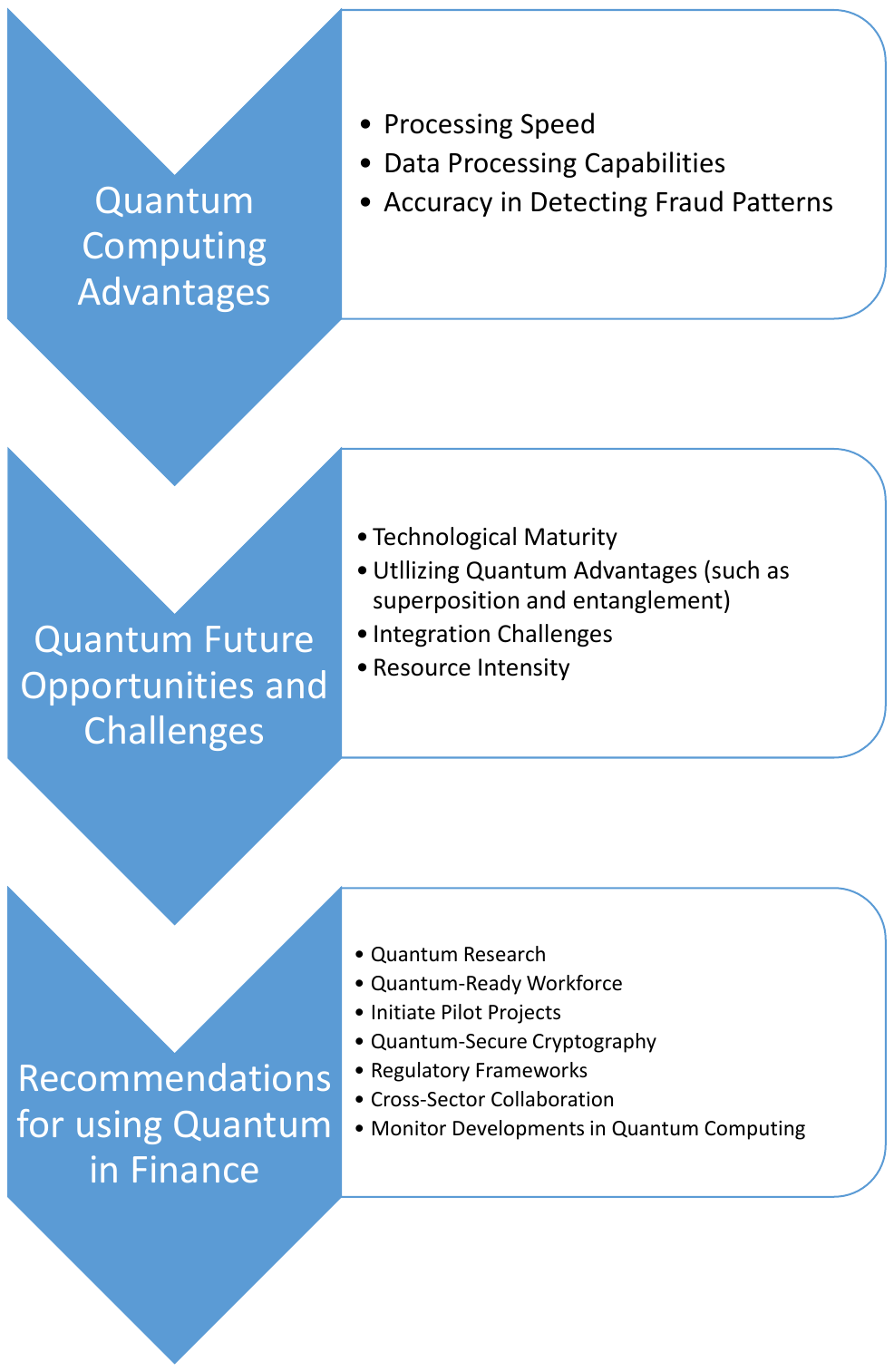}
	\end{center}
	\caption{Quantum Context: Advantages, Opportunities, and Challenges in Finance.}
	\label{fig:FigureQuantum}
\end{figure}

In comparing classical and quantum computing approaches for detecting financial crimes, the distinctions largely revolve around speed, data processing, and pattern detection accuracy. Classical computing, the foundation for fighting financial crime, processes information in binary, limiting its speed and efficiency as data volumes grow. As can be seen in Figure \ref{fig:FigureQuantum}, quantum computing, leveraging principles like superposition and entanglement, promises significant improvements in these areas but remains in the early development stages for practical application.

\begin{itemize}
    \item[-] Processing Speed: Classical methods are bound by algorithms’ linear or polynomial time complexities, which can slow searching through massive datasets. Using algorithms like Grover’s, Quantum computing could dramatically reduce search times, offering a quadratic speedup. Indeed, quantum computers could potentially complete tasks in hours or days that would take classical computers much longer. 
  \item[-] Data Processing Capabilities: Classical computing has advanced with big data technologies and ML, enabling the processing of large datasets to identify fraud patterns. However, the increasing volume and complexity of data present challenges. Using techniques like QPCA, Quantum computing could theoretically manage high-dimensional data more efficiently, taking advantage of quantum computing’s ability to operate in high-dimensional spaces.
  \item[-]Accuracy in Detecting Fraud Patterns: The accuracy of classical AI and ML models in identifying fraud can be affected by the complexity of financial systems. Quantum computing’s approach to processing multiple possibilities simultaneously could theoretically improve model accuracy by uncovering subtle correlations missed by classical methods. However, the practical benefits of quantum computing in enhancing detection accuracy are yet to be fully realised. The transition from classical to quantum computing in financial crime detection is expected to bring significant advancements but is contingent on overcoming technological challenges and developing new algorithms and security measures. The potential for quantum computing to transform the field is clear, yet much work remains to harness these capabilities fully.
\end{itemize}

Based on the comparison between classical and quantum computing in detecting financial crimes, several findings emerge, leading to recommendations for future research, development, and policy formulation: 
\begin{enumerate}[label=\alph*), ref=\alph*]
  \item  Technological Maturity, given that quantum computing is nascent, especially in practical applications for financial crime detection. While promising, the technology requires further development to realise its full potential;
  \item Quantum Advantage: The theoretical advantages of quantum computing—such as processing speed and the ability to handle complex, high-dimensional data—are clear. However, practical demonstrations of these advantages in real-world financial crime detection scenarios are limited;
  \item Integration Challenges: Integrating quantum computing into financial crime detection frameworks poses significant challenges, including the need for new algorithms, data security measures, and possibly regulatory changes;
  \item Resource Intensity: Quantum computing currently requires substantial resources, including specialised knowledge and quantum computing hardware, which are not widely accessible.
\end{enumerate}
The above analyses provided some useful recommendations:
\begin{enumerate}[label=\alph*), ref=\alph*]
  \item Investment in Research and Development: Governments, academia, and the private sector should invest in R\&D to advance quantum computing technology, focusing on developing practical applications for financial crime detection;
  \item Develop Quantum-Ready Workforce: Educational institutions and industry players should work together to develop training programs that prepare a new generation of professionals skilled in quantum computing and its applications in finance;
  \item Pilot Projects: In collaboration with quantum computing firms, financial institutions should initiate pilot projects to explore the application of quantum computing in detecting financial crimes. These projects can provide valuable insights into the technology’s practical benefits and limitations;
  \item Quantum-Secure Cryptography: With the advancement of quantum computing, existing encryption methods could become vulnerable. Developing quantum-secure cryptography is essential to protect financial data against potential quantum computing threats;
  \item Regulatory Frameworks: Regulatory bodies should begin considering the implications of quantum computing on financial systems. Developing guidelines and standards that address the use of quantum computing in finance will be crucial to ensuring security, privacy, and fair use;
  \item Cross-Sector Collaboration: Encourage collaboration across financial institutions, technology companies, regulatory bodies, and academic researchers to share knowledge, resources, and best practices in applying quantum computing;
  \item Monitor Developments in Quantum Computing: Financial institutions should continuously monitor advancements in quantum computing to stay ahead of potential threats and opportunities. It includes keeping abreast of developments in quantum algorithms that could revolutionise financial crime detection.
\end{enumerate}

\section{Conclusions}
The development of quantum computing introduces significant advancements in detecting and countering financial crimes. This technology, leveraging the unique capabilities of quantum mechanics, offers a new set of tools for analysing data and identifying illicit activities within vast datasets, doing so at a pace and level of complexity that classical computing can not match. The examples of quantum algorithms explored in this article, from QSVM for anomaly detection to quantum clustering for identifying hidden networks involved in money laundering, showcase the potential applications across various financial crime categories.
Quantum computing’s edge lies in its ability to process information through mechanisms such as superposition and entanglement, providing a leap in the speed and efficiency of data analysis. This ability is critical in the financial sector, where detecting fraudulent patterns swiftly can significantly disrupt criminal operations. By integrating quantum technologies with existing frameworks and strategies for financial crime prevention, there is a clear path towards enhancing the effectiveness of these measures.
The alignment of financial crime classifications with recognised standards from global entities, such as the FATF and the UNODC, emphasises the need for a unified strategy incorporating quantum computing. This approach improves existing methods and introduces new capabilities for uncovering and combating financial crimes.
The financial sector faces an opportunity to strengthen defences against criminal activities. It includes investment in research, development, and collaborative efforts to explore and apply quantum technologies. The adoption of quantum computing in financial crime detection signals a move towards more secure, transparent, and resilient financial systems equipped to counter the challenges posed by sophisticated criminal schemes. This article encourages the financial community to seriously consider the potential of quantum computing. We can expect a significant enhancement in the mechanisms for preventing and detecting financial crimes safeguarding the financial system against emerging threats by adopting this technology.

\bibliographystyle{unsrt}  
\bibliography{references}

\end{document}